\documentclass{article}

\usepackage{ijcai16}
\usepackage{graphicx}
\usepackage{url}
\usepackage{times}
\usepackage{amsmath,amssymb,array}
\usepackage{multirow}
\usepackage{color}
\usepackage{subfig}
\usepackage[boxed]{algorithm2e}
\usepackage{tikz}
\usetikzlibrary{arrows,calc,matrix,shapes,trees,positioning}
\usepackage{pgfplots}
\usepackage{pgfplotstable}
\usetikzlibrary{positioning,shapes,shadows,arrows}
\usetikzlibrary{patterns}
\usepackage{xcolor}

\newcommand{\fakeparagraph}[1]{\noindent\textbf{#1.}}

\newcommand{\KOGNAC}{\texttt{KOGNAC}\xspace}
\newcommand{\comment}[1]{}

\title{KOGNAC: Efficient Encoding of Large Knowledge Graphs}
\author{
Jacopo Urbani\textsuperscript{\normalsize a}, Sourav Dutta\textsuperscript{\normalsize b}, Sairam Gurajada\textsuperscript{\normalsize b}, and Gerhard Weikum\textsuperscript{\normalsize b} \\
\textsuperscript{\normalsize a}VU University Amsterdam, The Netherlands\\
\textsuperscript{\normalsize b}Max Planck Institute for Informatics, Germany\\
jacopo@cs.vu.nl,\{sdutta,gurajada,weikum\}@mpi-inf.mpg.de}

\begin{document}

\maketitle

\begin{abstract}
Many Web applications require efficient querying of large Knowledge Graphs (KGs).
We propose \texttt{KOGNAC}, a dictionary-encoding algorithm designed to improve
SPARQL querying with a judicious combination of statistical and semantic
techniques. In \texttt{KOGNAC}, frequent terms are detected with a frequency
approximation algorithm and encoded to maximise compression. Infrequent terms
are semantically grouped into ontological classes and encoded to increase data
locality. We evaluated \texttt{KOGNAC} in combination with
state-of-the-art RDF engines, and observed that it significantly improves SPARQL
querying on KGs with up to 1B edges.
\end{abstract}

\section{Introduction}
\label{sec:intro}

The advent of natural-language-based queries and entity-centric search has led
to the enormous growth and applicability of {\em Knowledge Graphs} (KG) to model
known relationships between entity-pairs. Large KGs have not only been built in
academic projects like DBpedia~\cite{dbpedia},but are also
used by leading organizations like Google, Microsoft, etc., to support
user-centric Internet services and mission-critical data analytics.

KGs are generally represented using the RDF data model~\cite{rdf}, in which the
KG corresponds to a finite set of subject-predicate-object (SPO) triples whose
terms can be URIs, blank nodes, or literal values~\cite{rdf}. Since many Web
applications rely on RDF-style KGs during their processing, efficient and
scalable querying on huge KGs with billions of RDF triples have necessitated
intelligent KG representation.

In concept, KGs can be managed using a variety of platforms, like RDF
engines~\cite{triplebit,triad,rdf3x}, relational stores~\cite{monetdb},
or graph database systems~\cite{neo4j}. In this context, the storage of RDF
terms in raw format is both space and process inefficient since these are
typically long strings. As such, all existing approaches {\em encode} the RDF
terms typically by mapping them to fix-length integer IDs, with the original
strings retrieved only during execution.

\fakeparagraph{Objectives} Modern KGs are typically queried using the W3C SPARQL
language~\cite{sparql}. Currently, the impact of different ID mappings on
advanced SPARQL operations (like query joins, index compression, etc.) is less
well studied. Ideally, the encoding of RDF terms into numerical IDs should: i)
Consider the {\em skew} in the term frequencies in the KG, and assign smaller
IDs to frequent terms in order to facilitate efficient down-stream compression
(by the storage engine). ii) For more advanced query access patterns,
particularly for join operations, {\em data locality} should be increased as
much as possible by the encoding. That is, terms that are often accessed
together should have close ID assignment in order to further reduce memory and
index access~\cite{harbi2015}. iii) It is often crucial to quickly load billions
of triples, for example, when a KG is required as background knowledge for new
analytic applications, or for append-only bulk update operations. Thus, the
encoding process should support parallelism as much as possible for better
scale-up.

\fakeparagraph{Problem Statement} Current RDF engines generally employ four
types of encodings: \emph{order or hash-based}, \emph{syntactic}, or based on
\emph{coordinates}. Order-based approaches assign consecutive IDs for new
incoming triples in the order of appearance. Hash-based procedures use term
hashes as IDs.  Syntactic encoding assigns IDs to terms based on their
lexicographic order.  Coordinate-based techniques stored the terms in special
data structures and use memory coordinates as IDs.

Interestingly, we observe that none of the existing approaches performs well
along all three dimensions of our desiderata. Assigning consecutive identifiers
leads to good compression, but does not improve locality for joins. Hash-based
encoding allows parallel loading, but has poor locality and is sub-optimal in
exploiting skew. The last two disadvantages are also shared by methods which
use memory coordinates as IDs. Syntactic encoding provides a compromise along
the three objectives, but is not robust enough to handle the cases where term
similarity cannot be extracted directly from the syntax. The problem addressed
in this paper is to encode RDF terms in large KGs such that all three
desiderata, namely better compression, query performance, and loading time, are
well satisfied.

\fakeparagraph{Contribution} We present \KOGNAC (\textbf{K}n\textbf{O}wledge
\textbf{G}raph e\textbf{N}oding \textbf{A}nd \textbf{C}ompression), an efficient
algorithm for KG encoding based on a judicious combination of statistical and
semantic techniques. Our encoding procedure detects {\em skewness} in term
frequency distribution with a approximation streaming technique, and subsequently
encodes frequent terms differently in order to facilitate high down-stream
compression. To improve {\em data locality} for join access patterns, \KOGNAC
computes semantic relatedness between terms by hierarchically grouping
them into ontological classes, and mapping terms in the same group to
consecutive IDs.

\KOGNAC has the advantage that it is independent from RDF application details,
since its output is a plain mapping from strings to IDs. To evaluate its
efficiency, we integrated it with four RDF systems -- RDF-3X~\cite{rdf3x},
TripleBit~\cite{triplebit}, MonetDB~\cite{monetdb}, and TriAD~\cite{triad} --
and observed significant improvements in query performance on metrics like
runtime, RAM usage, and disk I/O.

A longer version of this paper, with more details and experiments, is available
online at~\cite{kognacreport}.
 \section{Encoding KGs: State Of The Art}
\label{sec:problem}

Typically, applications query KGs using SPARQL~\cite{sparql} -- a W3C
declarative language. The core execution of SPARQL queries corresponds to
finding all graph isomorphisms between the KG and the graphs defined in the
queries.

\fakeparagraph{RDF Encoding} SPARQL engines, e.g., TripleBit~\cite{triplebit},
TriAD~\cite{triad}, Virtuoso~\cite{virtuoso}, etc., use {\em dictionary
encoding} to assign numeric IDs to terms based on their {\em appearance
ordering}, i.e., simply using consecutive or pseudo-random numbers for incoming
triples. The 4Store engine~\cite{4store} uses a string-hashing based ID
assignment that disregards any possible co-relation among terms. Both
approaches do not consider term frequencies leading to sub-optimal encoding with
frequent terms possibly assigned to larger IDs. Further, sophisticated
partitioning methods in TriAD renders such encoding prohibitively compute
expensive~\cite{harbi2015}.

RDF-3X~\cite{rdf3x}, one of the fastest single-machine RDF storage engines,
pre-sorts the SPO triples lexicographically and then assigns consecutive integer
IDs. A similar approach is also followed by~\cite{compr2013},
while~\cite{parallelencode} proposes a combination of appearance order with
hashing to improve partitioning. In contrast to our work, these approaches
strongly leverage the string similarity heuristics to cluster the elements.
These heuristics break when the semantic similarity does not follow the
lexicographic ordering. Such dissimilarity occurs frequently via subdomain usage
in URIs, or may even be imposed explicitly by political decisions (e.g.,
Wikidata~\cite{wikidata} uses meaningless strings to avoid an English bias).

Some relational engines (e.g., MonetDB~\cite{monetdb}) can optionally use
dedicated data structures for the storage of strings (mainly using variants of
Tries). In this context, a
particular variant of Trie with term prefix overlap was proposed
in~\cite{semanticmeta} to capture syntactic similarity. In these cases, the
coordinates to the data structure (e.g., memory addresses) are used as numerical
IDs. These IDs are typically long, and the induced locality reflects the
physical storage of the strings rather than the semantics in the KG. Older
versions of MonetDB followed this approach, but it was later abandoned.

\fakeparagraph{Semantic Relateness} There is a rich literature on semantic
relatedness based on the lexicographic features~\cite{zhang_recent_2013}, or on
domain-dependant data like Wikipedia~\cite{gabrilovich_computing_2007},
Wordnet~\cite{budanitsky_evaluating_2006}, biomedical
data~\cite{pedersen_measures_2007}, and spatial~\cite{seman}. In our case, we
cannot make assumptions about the domain of the input and the strings may be
completely random, so lexicographic features are not applicable.

In general, semantic relatedness functions cannot be directly applied to our
problem of graph encoding. For instance, \cite{semanticrel} defines semantic
relatedness among two nodes as a function of combining the path length and the
number of different paths between two nodes. In our context, it would be too
expensive to compute relatedness for many (or even any possible)
pairs of nodes. Furthermore, the high sparsity in the graph results in very low
relatedness coefficients in almost all cases. \cite{litemat} describes how
ontological taxonomies can be exploited to speed up reasoning via intelligent ID
encoding. In spirit, this approach is similar to our approach for encoding
infrequent terms. However, ~\cite{litemat} focuses on improving reasoning
efficiency rather than the semantic relatedness. Furthermore,~\cite{litemat}
does not consider data skewness, as we do.

Finally, clustering methods based on the graph structure (e.g.,
METIS~\cite{metis}, or graph-coloring approach of~\cite{rdf2db}) are infeasible
at our scale~\cite{triad}, and often require a conversion to an undirected
single-label graph disregarding entirely the semantics in the KG. In
contrast, the goal of \texttt{KOGNAC} is to leverage precisely this semantics to
improve the encoding.
 \section{The {\bf \Large \texttt{KOGNAC}} Algorithm}
\label{sec:overview}

\begin{figure}[tb]
\centering
    \includegraphics[width=0.8\linewidth]{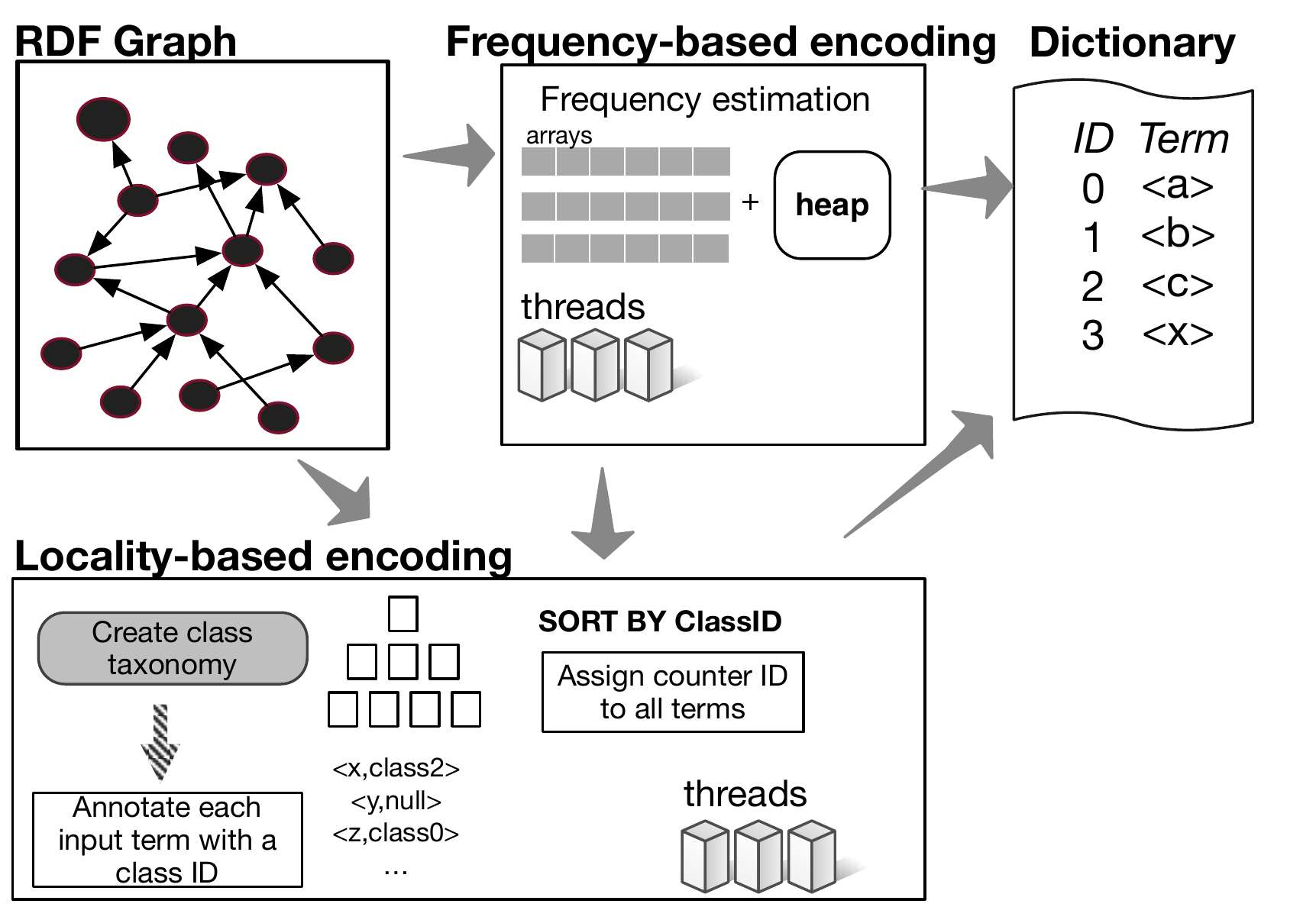}
    \caption{High level overview of KOGNAC.}
    \label{fig:overview}
\end{figure}


%

\fakeparagraph{Algorithmic Overview} \KOGNAC performs a crucial distinction
between (few) \emph{frequent} terms from the (many) remaining \emph{infrequent
ones}. For the frequent terms, it applies a frequency-based encoding, which is
highly effective in terms of compression due to the skewed distribution of
modern KGs~\cite{skewness}. For the
infrequent ones, it exploits the semantics contained in the KG to encode similar
terms together to improve data locality.

Fig.~\ref{fig:overview} describes at a high level the functioning of
\KOGNAC. Let $V$ be the set of terms in a generic RDF graph $G$. \KOGNAC
receives $G$ and a threshold value $k$ (used for the top-$k$ frequent elements) as
input, and returns a dictionary table $D \subset V \times \mathbb{N}$ that maps
every element in $V$ to a unique ID in $\mathbb{N}$.

The dictionary table $D$ is constructed using two different encoding algorithms:
\emph{Frequency-based encoding (FBE)}, which encodes only the frequent terms,
and \emph{Locality-based encoding (LBE)}, which encodes the infrequent ones. The
core computation of FBE corresponds to the execution of an approximated
procedure for accurate frequency detection. LBE instead constructs a class
taxonomy, groups the terms into these classes, and assigns the IDs accordingly.
Both methods support parallelism. FBE is executed before LBE. If $D_1$ is the
output of FBE, and $D_2$ of LBE, then $D=D_1 \cup D_2$ and $D_1 \cap D_2 =
\emptyset$.
 \section{Frequency-based Encoding (FBE)}
\label{sec:sampling}

The goal of this procedure is to detect the top-$k$ frequent terms and assign
them incremental IDs starting from the most to the least frequent term. For our
purpose, an exact calculation of the frequencies is not required, and even
though it can be easily computed for small KGs, it would be unnecessarily
expensive for very large KGs. Sampling provides the reference technique for a
fast approximation~\cite{compr2013}.  Unfortunately, an excessive sampling might
lead to false positives and negatives, and increasing the sample size for
tolerable error rates might still be too expensive.

To obtain a better approximation, we investigated the applicability of
hash-based sketch algorithms~\cite{frequentvalues,countsketch}. Sketch
algorithms have successfully been deployed in other domains to identify
distinct items in streams~\cite{frequentvalues}, but never to our problem
domain.

\fakeparagraph{Sketch Algorithms} We experimented with three state-of-the-art
sketch algorithms: {\em Count-Sketch}~\cite{countsketch}, {\em
Misra-Gries}~\cite{misragries}, and {\em Count-Min}~\cite{countmin}.
Count-Sketch, a single-scan algorithm, requires a heap with quadratic
space in error tolerability. After many experiments, we concluded that updating
such large heap was too expensive for our inputs.

Misra-Gries is similar to Count-Sketch, with the difference that it uses a
smaller heap and reports the terms that are at least {\em k-frequent}, i.e.,
having a frequency $> \frac{n}{k}$, where $n$ denotes the total number of term
occurrences. This method approximates the relative frequencies (i.e., frequency
after a term is inserted in the heap). We found that the relative frequencies
were not accurate enough to allow a precise ordering of terms, as the count
depends on the appearance ordering.

Count-Min uses $n>1$ hash counter arrays and $n$ hash functions to count the
frequencies. In contrast to the previous two methods, it always requires two
input scans: First to count the frequencies, and then to extract the actual
frequent terms. Count-Min does not use heaps and this makes it in principle
faster than the other two. However, for large inputs the cost of second input
scan offsets this gain.

\begin{figure}[t] \centering
    \includegraphics[width=0.9\linewidth]{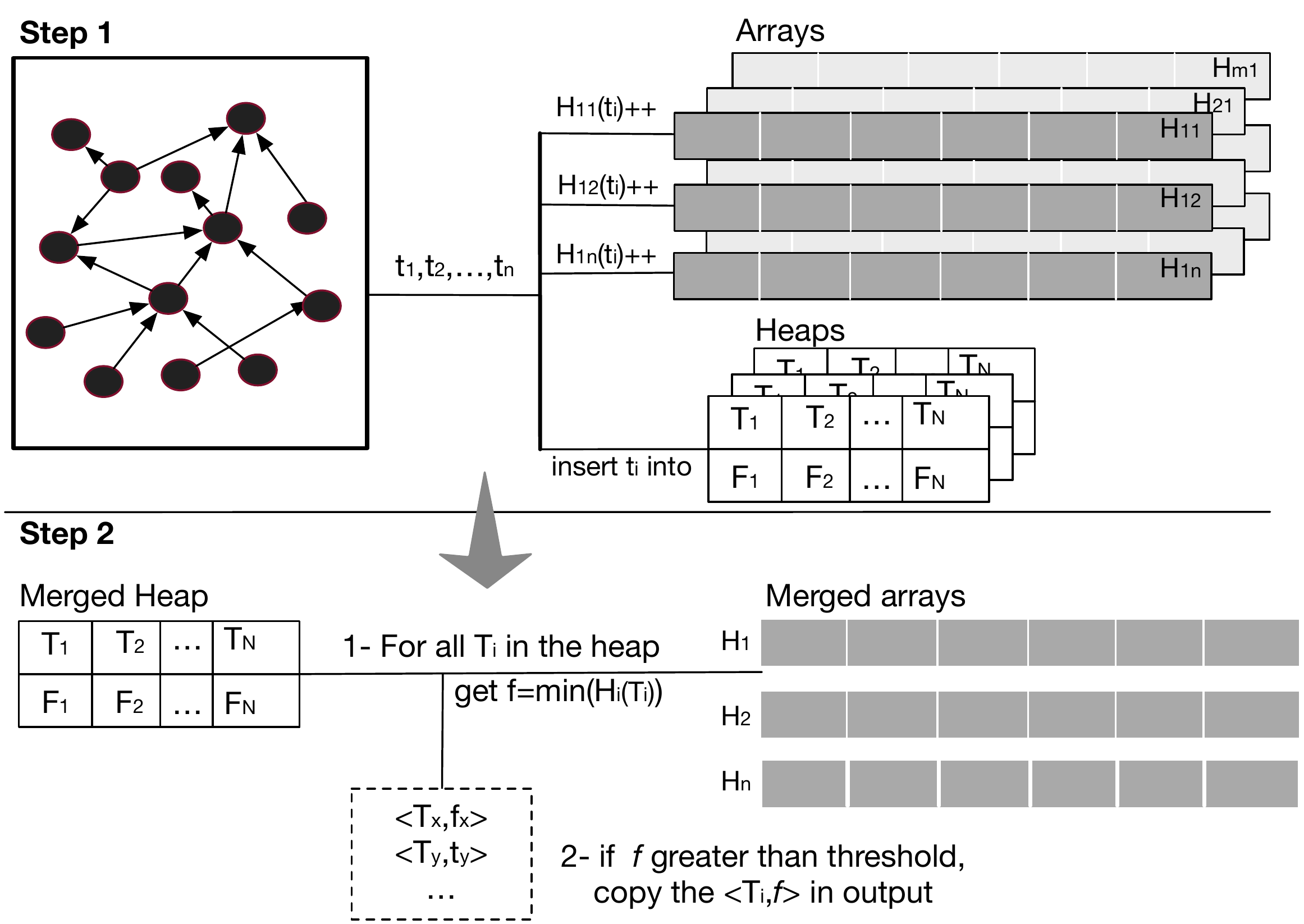} \caption{Overview of
    CM+MG.} \label{fig:cmmg}
\end{figure}

\fakeparagraph{Our proposal: CM+MG} Count-Min provides a good estimate of term
frequencies, but cannot identify the top-k elements within a single pass.
Misra-Gries detects the top-k elements but does not report good frequencies. The
disadvantages of the two are complementary. We thus propose a hybrid approach,
which we call {\em CountMin+MisraGries (CM+MG)}, that intelligently combines
elements of the two.

An overview of its functioning is reported in Fig.~\ref{fig:cmmg}. As input it
receives the input KG, $k$, a hash family $H$ with $n$ hash functions, $m$
parallel threads, and a threshold $k$ of popular elements. In our
implementation, we selected as default values three hash functions for $H$,
while $m$ is the number of physical cores, and the value $k$ is requested from
the user.

CM+MG is executed in two steps: In the first step, CM+MG creates $m*n$ counter
arrays and $m$ Misra-Gries heaps of size $k$. The KG is split into $m$ subsets,
and fed to the $m$ threads. Each thread calculates $n$ hash codes for each term
occurrence in its partition. The $n$ hash-codes are modulo-ed the array size and
the corresponding indices in the arrays are incremented by $1$. The terms are
also inserted into the heaps.

In the second step, the $m$ heaps are merged in a single heap.
Also the arrays are summed into $n$ final arrays. As
threshold value for the frequent terms, we select the top-k value in the first
array. The algorithm now scans all elements in the merged heap. Instead of using
the relative frequency as estimate, CM+MG queries the arrays using the term's
hashcodes, and uses the minimum of the returned values. If this value is greater
than the threshold, the term is marked as frequent.

\fakeparagraph{Effectiveness of FBE} Our approach works well because KGs are
skewed. In order to better characterize the gain we can obtain with our
approach, we present a short theoretical analysis on the space efficiency of
FBE.

Assume $T=\{t_1, t_2, \ldots, t_n\}$ to be $n$ distinct RDF terms,
with term $t_i$ having a frequency of $f_i$ in the input KG. In the worst case,
an assignment criterion which is independent from the term frequencies (e.g., an
order-based one) will produce an assignment which is close to a fixed-length
encoding; that is where all terms will be assigned to IDs of length $\lceil
\log_2{n} \rceil$ bits. In this case, the total space required to store an
encoded KG in the database would be
{\small{
\begin{align} \label{eq:fix} S_{fix} = \sum_{i=1}^n f_i\lceil \log_2{n} \rceil =
    F\lceil \log_2{n} \rceil \quad \text{[with $F = \sum_{i=1}^n f_i$]}
\end{align}
}}
\normalsize

In \KOGNAC, the terms are divided into blocks depending on their
frequencies. Here, block $i \in \{1\ldots b\}$ contains the top $2^{i}$ elements
which are not in any previous group. Hence, there exists $b=\lceil \log_2{n}
\rceil$ non-empty blocks for $n$ distinct items in $T$. An item in block $i$ is
encoded using $i$ bits. Since the assignment is not \emph{prefix-free}, in
order to properly decode the IDs we need to append to each term some extra
data to discriminate it from the different values. This extra data must
take at least $\lceil \log_2{b} \rceil$ bits as this is the minimum space necessary
to identify each of the $b$ blocks. Hence, the total space
required for decoding an item in block $i$ takes $(i+\lceil \log_2{b} \rceil)$
bits. Assuming, $f_j^i$ to denote the frequency of the $j^{th}$ item in block
$i$, the total encoding space required for KG is,
\small
\begin{align}
\label{eq:kog1}
    S_{kog} &= \lceil \log_2{b} \rceil \sum_{i=1}^b
    \sum_{j=1}^{2^i} f_j^i + \sum_{i=1}^b i \sum_{j=1}^{2^i} f_j^i \end{align}
\normalsize

Since modern KGs have a skewed term distribution~\cite{skewness}, we now assume
that the item frequencies is drawn from a Zipfian distribution\footnote{This
    distribution is used for heavy-tailed characteristics observed in natural
    language sources used for KG construction.} with parameter $s \geq 2$, such
    that the frequency of the $k^{th}$ frequent term $f_k \approx
    \frac{F}{s^k}$.

\noindent If we apply this distribution in Eq.~\eqref{eq:kog1}, we obtain

\begin{align} S_{kog} &= F
    \lceil \log_2{b} \rceil + \sum_{i=1}^b \frac{iF}{s^{\sum_{k=1}^{i-1} 2^k}}
    \sum_{j=1}^{2^i} \frac{1}{s^j} \nonumber \end{align}

By algebraic manipulations, we have

\begin{align} \label{eq:kog2} &S_{kog} = F \lceil
    \log_2{b} \rceil + \sum_{i=1}^b \frac{iF}{s^{2^i-2}} . \frac{1}{s-1} \qquad
    \text{[for large $i$]} \nonumber \\ &\approx F \lceil \log_2{b} \rceil + F
    \sum_{i=1}^b \frac{i}{s^{2^i}} \approx F \left(\lceil \log_2{b} \rceil +
    \frac{1}{s^2}\right) \end{align}
If we compare Eq.~\ref{eq:kog2} with Eq.~\ref{eq:kog1}, then we see that with
\KOGNAC we can achieve nearly an exponential theoretical decrease (i.e.,
$log~log~{n}$ vs. $log~{n}$) in the total encoding space required to store the
KG. This is the scale of potential improvement that our encoding can offer to
the current frequency-independent encoding algorithms.
 \section{Locality-based Encoding (LBE)}
\label{sec:semantic}

\begin{figure}[tb]
    \includegraphics[width=\linewidth]{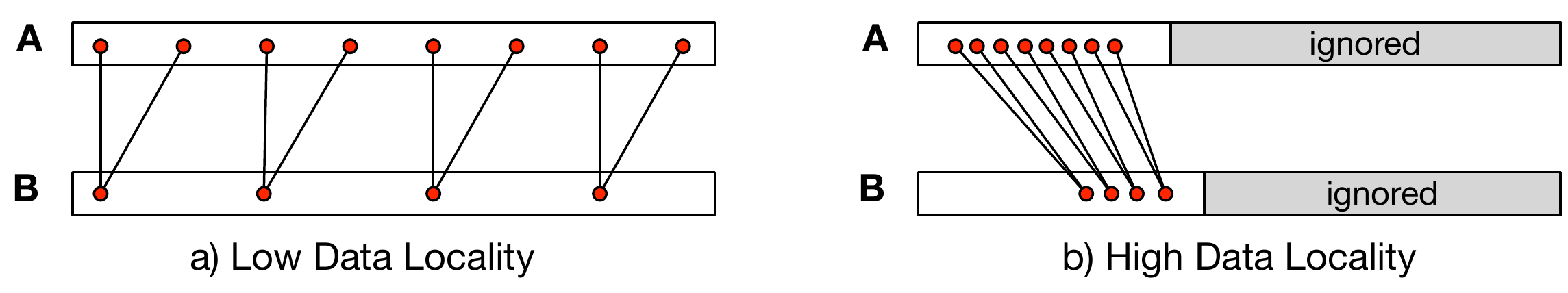}
    \caption{Effect of data locality during SPARQL join.}
    \label{fig:datalocality}
\end{figure}

\begin{algorithm}[t]
    \footnotesize
    \caption{Locality-aware Encoding: \emph{Input:} a KG, the taxonomy $T$,
    and the frequent dictionary $D_{freq}$ generated by FBE. \emph{Output:} The infrequent
    dictionary $D_{infreq}$.}
    $S$ := \{\}; $D_{infreq}$ := \{\}; $cID$ := max ID in $D_{freq}$\;
 $MAX$ := constant with number higher than any class ID in $T$\;
 \For{every triple $<$s,p,o$>$ in the KG }{
  Add to $S$ three pairs: $\langle s,MAX \rangle$,$\langle p,MAX \rangle$,$\langle o,MAX \rangle$\;
  If $p$ = 'type', then add $\langle s, id(o,T)\rangle$ to $S$\;
  If $p$ has domain $c$, then add $\langle s,id(c,T)\rangle$ to $S$\;
  If $p$ has range $c$, then add $\langle o,id(c,T)\rangle$ to $S$\;
}
Remove from $S$ all pairs $\langle t,c \rangle$ where $t$ is in $D_{freq}$\;
\For{all pairs $\langle t_1,c_1 \rangle$ and $\langle t_2,c_2 \rangle$ in $S$ }{
    \If{$t_1 = t_2$}{
        Remove $\langle t_1,c_1 \rangle$ from $S$ if $c_1>c_2$ or remove $\langle t_2,c_2
        \rangle$ otherwise\;
    }
}
\While{$S$ is not empty}{
    Take out from $S$ one pair $\langle t_1,c_1\rangle$ s.t. $\nexists \langle t_2,c_2\rangle \in S : c_2 < c_1 \vee c_1 = c_2 \wedge t_2 < t_1 $\;
    Increment $cID$ by 1\;
    Add to dictionary $D_{infreq}$ the assignment $\langle t_1,cID\rangle$\;
}
\label{alg:1}
\end{algorithm}

In the long tail of the frequency distribution, a frequency-based encoding no
longer pays off. Each infrequent term appears only a few times and this reduces
the negative impact of assigning large IDs to them. Moreover, the increased ID space
provides a much larger number of disposable IDs: for instance, after the most
frequent $2^{24}$th ID, all the following $2^{32}-2^{24}$ IDs will take the same
number of bytes.

\fakeparagraph{Data Locality} With \texttt{LBE}, we propose an encoding that is
designed to improve \emph{data locality} during the execution of SPARQL queries.
Data locality plays a significant role to reduce the cost of
index access for advanced operations like relational joins. Consider, as
example, Fig.~\ref{fig:datalocality}, which shows a join between two generic
relations $A$ and $B$. It is common that these relations are indexed (i.e.,
sorted) to enable merge joins~\cite{rdf3x}. If the index locations of the join
terms are spread around the entire relations, as shown in
Fig.~\ref{fig:datalocality}a, then the join algorithm must process large parts
of the indexes. On the contrary, if the join succeeds using only sub-portions of the
index, then the join algorithm can save significant computation by ignoring
large chunks of the indexes (Fig.~\ref{fig:datalocality}b).

\fakeparagraph{Encoding and Data-Locality} Since SPARQL engines work (mostly)
directly on the encoded data, the elements on which the join operates are
precisely the IDs which we should assign during encoding. Therefore, an
assignment of close IDs will significantly improve data locality.

Unfortunately, it is not possible to encode the terms so that they are
\emph{always} next to each other. We can make one assignment only, and SPARQL
queries could request joins on any subset of the relations. Still, we can
leverage a heuristics that is surprisingly effective: SPARQL joins tend to
materialize between terms which are semantically related.

Following this heuristic, we propose to cluster the terms into the ontological
classes they are connected to via the \texttt{isA} relation, and assign
consecutive IDs to the members of each cluster. Our approach has several
advantages: (i) \texttt{isA} is a common relation in KGs and is
domain-independent; (ii) new \texttt{isA} edges can be inferred using other
ontological information, e.g., definitions of the domain/range of properties;
(iii) classes can be organized in a taxonomy using the \emph{rdfs:subClassOf}
relation of the RDF Schema~\cite{rdfs}. The taxonomy can help us to further
separate instances of subclasses from instances of siblings of the parents
(e.g., \emph{Students} should be closer to \emph{Professors} than to
\emph{Robots} because the first two are both subclasses of \emph{Persons}).

\fakeparagraph{Algorithm Overview} Our locality-based encoding works as follows:
First, we must create the taxonomy of classes. To this end, we extract all
triples that define the subclass relation between classes, or mention classes (these share
\emph{rdfs:subClassOf} as predicate, and are the objects of triples with the
predicates \emph{isA}, \emph{rdfs:domain}, and \emph{rdfs:range}). We create a
graph where the classes are vertices and the edges are defined by the
\emph{subclassOf} triples. We add the standard class \emph{rdfs:Class} (the
class of all classes~\cite{rdfs}), and add one edge from each vertex to it, to
ensure that there are no disconnected components. We remove possible loops in
the graph by extracting the tree with the maximum number of edges rooted in
\emph{rdfs:Class}.

We now assign in post-order an incremental class ID to each class in the tree.
We indicate with $id(c,T)$ the class ID assigned to the class $c$ contained in
the taxonomy $T$.

After the taxonomy is built, we are ready to encode the terms. This procedure is
outlined in Alg.~\ref{alg:1}. First, we annotate each term with a class ID it is
an instance of. The annotation might come from an explicit relation
(\texttt{isA}) or an implicit one (\texttt{domain} and \texttt{range}). If a
term cannot be associated to any class, we give it a dummy ID ($MAX$). Then, for
each term we maintain only the annotation with the class which has the smallest
ID. Finally, we order all annotations first by class ID, and then
(lexicographically) by term. We use the same counter used in FBE and assign
incremental IDs to the terms with the order defined in the sorted list. In this
way, the assignment first considers the semantic type of the term, and then its
syntax. Notice that terms which are not mapped to any class (mainly labels),
will be encoded only syntactically.

A large part of the computation of LBE can be parallelized. The task of
assigning the terms to the smallest class IDs does not require thread
synchronization because at this point the taxonomy is a read-only data
structure. Therefore, the task can be trivially parallelized using standard
input range partitioning and parallel merge sorts. The final assignment is
performed sequentially due to the usage of a single counter.
 \section{Evaluation}
\label{sec:evaluation}

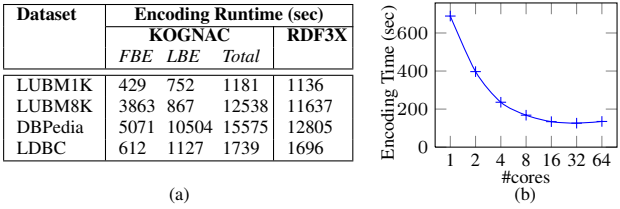
\begin{figure}[t]
    \begin{tikzpicture}
    \node at (0,0) {
    \begin{minipage}{0.5\linewidth}
    \scriptsize
      \tabcolsep=0.17cm
        \begin{tabular}{|p{1cm}|p{0.3cm}p{0.4cm}p{0.5cm}|p{0.7cm}|}
            \hline
            \bf Dataset & \multicolumn{4}{c|}{\bf Encoding Runtime (sec)} \\
            \cline{2-5}

            & \multicolumn{3}{c|}{\bf KOGNAC} & \bf RDF3X \\
            & {\it FBE} & {\it LBE} & {\it Total} &  \\
            \hline
            \hline
            LUBM1K & 429 & 752 & 1181 & 1136 \\
            LUBM8K & 3863 & 867 & 12538 & 11637 \\
            DBPedia & 5071 & 10504 & 15575 & 12805 \\
            LDBC & 612 & 1127 & 1739 & 1696\\
            \hline
        \end{tabular}
    \end{minipage}
    };
    \scriptsize
 \begin{axis}[at={(0.42\linewidth,-0.1\linewidth)}, xmode=log, ymin = 0, ylabel= {Encoding Time (sec)}, xlabel = {\#cores}, scaled ticks=true, legend style={legend columns = -1, at={(1.95,1.2)},anchor=east,}, width=4cm, height=3.5cm, every axis x label/.style={
    at={(0.5,-0.3)},
    anchor=south,
}, every axis y label/.style={
    at={(-0.25,1.03)},
    anchor=east, rotate=90,
}, xtick={1,2,4,8,16,32,64}, xticklabels={1,2,4,8,16,32,64}]
\addplot[smooth,mark=+,blue] plot coordinates{
(1,689)
(2,397)
(4,236)
(8,168)
(16,134)
(32,126)
(64,135)

};
\end{axis}
    \node at (0.2,-1.5) {(a)};
    \node at (4.8,-1.5) {(b)};
    \end{tikzpicture}
    \caption{(a) Comparison of KOGNAC vs RDF-3X dictionary encoding times, (b)
    KOGNAC multi-threading performance analysis on LUBM1K. Experiments on machine
M1.}
    \label{fig:enctimes}
\end{figure}

We implemented {\tt KOGNAC} in a C++ prototype. We tested it in combination with
four SPARQL engines: RDF-3X, TripleBit, TriAD, and MonetDB. We chose them
because they represent the state-of-the-art of different type of SPARQL engines:
native/centralized (RDF-3X, TripleBit), native/distributed (TriAD), and
RDBMS/centralized (MonetDB). These engines also perform different encodings:
RDF-3X performs syntactic encoding, TripleBit and TriAD performs an order-based
encoding, MonetDB can be loaded with arbitrary encodings (we used a syntactic
encoding).

We used two types of machines: \emph{M1}, a dual 8-core 2.4 GHz Intel CPU, 64 GB
RAM, and two disks of 4 TB in RAID-0; and \emph{M2}, a 16 quad-core Intel Xeon
CPUs of 2.4GHz with 48GB of RAM. As input, we used three RDF graphs in NT
format: LUBM~\cite{lubm} -- a popular benchmark tool, LDBC~\cite{ldbc}, another,
more recent benchmark designed for advanced SPARQL 1.1 workloads, and
DBPedia~\cite{dbpedia}, one of the most popular KGs. We created two LUBM
datasets: \emph{LUBM1K} (133M triples, 33M terms), and \emph{LUBM8K} (1B
triples, 263M terms). We created a LDBC dataset with 168M triples and 177M
terms. The DBPedia dataset contains about 1B triples and 232M terms.

For LUBM, we used five adaptations of benchmark queries which were selected as
representative in~\cite{triplebit}. For DBPedia, we slightly changed five
example queries that are reported in the project's website. For LDBC, we use the
official queries in the SPB usecase~\cite{ldbc}. These queries require SPARQL
1.1 operators which were unsupported by any of our engines. We implemented some
of the missing operators in RDF-3X so that we could launch 7 of the 12 SPB
queries. Fig.~\ref{fig:queries} reports the LUBM and DBPedia queries in
compressed form.  The LDBC queries are freely available at benchmark website
\url{www.ldbcouncil.org}. Here we use abbreviations to refer to them (e.g. query
Q3 is the third query of the official benchmark). The \texttt{KOGNAC} code is
available at \texttt{https://github.com/jrbn/kognac}.

\fakeparagraph{Encoding runtime} As baseline, we selected the syntactic encoding
algorithm performed by RDF-3X. We found that syntactic encoding performs better
than the others, and RDF-3X implements a highly optimized loading procedure.

First, we measured the (sequential) encoding runtimes of all four datasets and
report them in Fig.~\ref{fig:enctimes}(a). We notice that our approach is
slightly slower than the one of RDF-3X. We expected such difference, since we
perform a much more complex operation than a simple syntactic encoding. In the
worst case, \texttt{KOGNAC} is about 20\% slower. Considering that loading is a
one-time operation whose cost gets amortized over time, we deem it as an
acceptable cost, especially in view of the benefit we get at query time. In
Fig.~\ref{fig:enctimes}(b), we measured the \texttt{KOGNAC} runtime doubling
each time the number of threads from 1 to 64. We see that the runtime steadily
decreases until it stabilizes after 8 threads. The runtime left after this point
is the one necessary to sequentially assign the IDs to the list of terms and
write the dictionary to disk.

In a series of experiments (not shown in this paper), we compared the
performance of CM+MG against sampling at 5\% (the most popular technique) and
Count-Min (the fastest one). We varied the $k$ threshold, and measured the
runtime and accuracy of the approximation. We found that $k=50$ was a good
threshold value because KGs typically contain only few very frequent terms.
Therefore, all experiments in this section should be intended with $k=50$. In
general, CM+MG was the fastest algorithm with $k$ up to $500$. In the best case,
it was twice as fast as Count-Min (the second best). In the worst case, it was
10\% slower.  With higher $k$s, CM+MG became slower because of the heap and
Count-Min returns the best runtimes. In terms of accuracy, all three managed to
identify the very first top $k$. However, as we increased $k$, all methods
started to fail: Sampling quickly lost accuracy, Count-Min produced large
overestimates (due to hash collisions in the arrays), while CM+MG produced
underestimates (due to the limited heap).

\begin{table*}
    \centering
    \scriptsize
    \tabcolsep=0.17cm
    \begin{tabular}{|p{1mm}|p{2.8mm}|p{7.7mm}|rr|rr|rr|rrr|}
            \hline
            & \bf Q. & \bf\mbox{\# Results} & \multicolumn{2}{c|}{\bf Cold runtime
            (sec)} & \multicolumn{2}{c|}{\bf Max RAM (MB)}&
            \multicolumn{2}{c|}{\bf Disk I/O (MB)} &\multicolumn{3}{c|}{\bf Avg.
                warm
            runtimes (sec) and S.T.}\\
            &&&{\it KOG}&{\it R3X}&{\it KOG}&{\it R3X}&{\it KOG}&{\it R3X} &{\it KOG}&{\it R3X}&{\it SS(p-value)}\\
            \hline
            \hline
\parbox[t]{1mm}{\multirow{5}{*}{\rotatebox[origin=c]{90}{\bf
            LUBM1K}}} & L1 & 10 &  0.22  &  0.31  &  4  &  4  &  14  &  19 & 0.002&0.004&2.97e-06\\
 & L2 & 10 &  0.04  &  0.17  &  5  &  6  &  16  &  27& 0.003&0.011&9.64e-05\\
 & L3 & 1 &  88.53  &  90.83  &  708  &  854  &  53  &  69&87.688&90.745&4.52e-04 \\
 & L4 & 2528 &  92.21  &  98.41  &  724  &  883  &  548  &  367 &87.365&89.774&4.33e-03\\
 & L5 & 44190 &  7.45  &  15.79  &  1,261  &  1,588  &  1,102  &  2,132&2.495&7.448&1.12e-07 \\
            \hline
\parbox[t]{1mm}{\multirow{5}{*}{\rotatebox[origin=c]{90}{\bf
            LUBM8K}}} & L1 & 10 &  0.09  &  0.27  &  4  &  4  &  15  &  21&0.002&0.004&1.12e-05 \\
 & L2 & 10 &  0.02  &  0.64  &  5  &  7  &  18  &  32&0.004&0.017&1.79e-04\\
 & L3 & 1 &  700.58  &  716.55  &  5,335  &  6,537  &  309  &  486 &699.219&711.499&1.28e-08 \\
 & L4 & 2528 &  717.65  &  744.10  &  5,320  &  6,536  &  321  &  811 &699.750&714.031&2.69e-07 \\
 & L5 & 351919 &  75.90  &  174.16  &  9,739  &  12,400  &  8,832  &  16,928&19.067	& 62.626&7.56e-11 \\
            \hline
\parbox[t]{1mm}{\multirow{5}{*}{\rotatebox[origin=c]{90}{\bf
            DBPedia}}} & D1 & 449 &  0.99  &  3.32  &  8  &  15  &  55  &  52 & 0.016	& 0.033	&4.53e-03\\
 & D2 & 600 &  0.23  &  3.17  &  6  &  8  &  29  &  61 &0.003	& 0.003 & 8.91e-01 \\
 & D3 & 270 &  1.60  &  2.96  &  6  &  11  &  59  &  49 & 0.005 & 0.018	& 6.59e-04\\
 & D4 & 68 &  0.72  &  1.34  &  6  &  6  &  45  &  49 & 0.011	& 0.009 & 3.88e-02 \\
 & D5 & 1643 &  5.05  &  26.79  &  29  &  60  &  330  &  263 & 0.031	& 0.049	& 2.13e-02\\
            \hline
\parbox[t]{1mm}{\multirow{7}{*}{\rotatebox[origin=c]{90}{\bf
            LDBC}}} & Q2 & 36 &  44.59  &  45.58  &  1,320  &  2,053  &  241  &  279  & 41.542&47.024 & 9.84e-02\\
 & Q3 & 178 &  126.48  &  132.55  &  524  &  574  &  588  &  624 & 122.342&130.517&3.95e-07\\
 & Q6 & 3819127 &  60.17  &  71.32  &  2,157  &  5,198  &  1,268  &  3,953 &46.879	& 52.918&2.52e-05 \\
 & Q7 & 98 &  5.85  &  6.76  &  549  &  3,663  &  625  &  3,675 &3.454&4.286&1.86e-07 \\
 & Q8 & 1018 &  1,847  &  4,915  &  2,867  &  5,934  &  1,804  &  3,949 & 26.613 &	31.002 & 3.68e-04\\
 & Q10 & 14 &  420.88  &  4,577  &  714  &  3,662  &  729  &  3,676 & 5.942&5.745&3.68e-03\\
 & Q11 & 114 &  24.51  &  89.21  &  170  &  204  &  201  &  237 & 1.163& 1.194&7.02e-01 \\
            \hline
        \end{tabular}
        \caption{Query runtime, Max RAM usage, and disk I/O with KOGNAC and RDF-3X
        encodings on one M1 machine.}
        \label{tab:queries}
\end{table*}

\fakeparagraph{SPARQL Query performance} Tab.~\ref{tab:queries} reports on the
execution of SPARQL queries with the mappings produced by \texttt{KOGNAC} and by
RDF-3X's syntactic encoding. The table reports cold query runtimes, the maximum
amount of RAM used by the system, and the disk I/O. The last three columns
report the warm runtimes, calculated as the average of five subsequent
executions, and a significance test calculated from them. From the table, we see
that \texttt{KOGNAC} leads to a significant improvement over all metrics.  All
cold query runtimes and all but two warm runtimes are faster, with improvements
of up to ten times. The system always uses less main memory and in all but four
cases it reads less data from disk.

We tested \texttt{KOGNAC} also with the other three systems, which use a
traditional order-based encoding. For conciseness, we report in
Tab.~\ref{tab:othersys} only the runtimes of the LUBM queries as they are
representatives of the general behaviour. We observe that also here
\texttt{KOGNAC} produced better runtimes. For TriAD, there was an improvement
in all but two cases, where the runtime was unchanged. For TripleBit, one
query failed due to bugs in the system, while another produced a slightly worse
runtime. In the remaining cases \texttt{KOGNAC} encodings were beneficial.
Finally, with MonetDB we observed an improvement in all cases. In essence, our
results show that an intelligent assignment, like the one produced by
\texttt{KOGNAC}, has a significant impact on the processing of the
KG. Given the encoding runtime, this improvement comes at little cost.

\begin{table}
   \scriptsize
    \tabcolsep=0.18cm
    \begin{tabular}{|l|rr|rr|rr|}
        \hline
        {\bf Q.} & \multicolumn{2}{c|}{\bf TriAD} &
        \multicolumn{2}{c|}{\bf TripleBit}& \multicolumn{2}{c|}{\bf
        MonetDB} \\
        &{\it KOGNAC}&{\it Standard}&{\it KOGNAC}&{\it Standard}&{\it
        KOGNAC}&{\it Syntactic}\\
        \hline
        \hline
        L1&0.001&0.001&0.056&0.149&0.820&2.4\\
        L2&0.002&0.002&0.094&n/a&0.943&1.2\\
        L3&0.106&0.631&1.672&1.567&11.1 & 15.2\\
        L4&2.684&3.090&5.626&6.549&9.5&21.1\\
        L5&2.558&3.067&5.082&6.438&4.1&8.2\\
        \hline
        \end{tabular}
        \caption{Impact of \texttt{KOGNAC} on example SPARQL queries using one M2 machine.}
        \label{tab:othersys}
\end{table}

\begin{figure}[t]
    \fbox{
    \begin{minipage}[t]{8cm}
\scriptsize
\fakeparagraph{LUBM Queries}\\
{
\noindent @prefix r: $<$http://www.w3.org/1999/02/22-rdf-syntax-ns\#$>$ \\
@prefix u: $<$http://www.lehigh.edu/$\sim$zhp2/2004/0401/univ-bench.owl\#$>$

\noindent {\bf L1.} \{ ?x u:subOrganizationOf $<$ http://www.Department0.
University0.edu$>$ .\\ ?x r:type u:ResearchGroup .\}

\noindent {\bf L2.} \{ ?x u:worksFor $<$http://www.Department0.University\-0.edu$>$ . ?x r:type u:FullProfessor . ?x u:name ?y1 . ?x u:emailAddress ?y2 . ?x u:telephone ?y3 . \}

\noindent {\bf L3.} \{ ?y r:type ub:University . ?x u:memberOf ?z . ?z u:subOrgOf ?y . ?z
r:type u:Department . ?x u:undergradDegreeFrom ?y . ?x
r:type u:UndergradStudent.\}

\noindent {\bf L4.} \{ ?y r:type u:University . ?z u:subOrgOf ?y . ?z r:type u:Department . ?x u:memberOf ?z . ?x r:type u:GraduateStudent . ?x u:undergradDegreeFrom ?y .\}

\noindent {\bf L5.} \{ ?y r:type u:FullProfessor . ?y u:teacherOf ?z . ?z r:type u:Course . ?x u:advisor ?y . ?x u:takesCourse ?z . \}
}

\fakeparagraph{DBPedia Queries} \\
{
\noindent @prefix foaf: $<$http://xmlns.com/foaf/0.1/$>$, purl: $<$http://purl.org/dc/terms/$>$, db:
$<$http://dbpedia.org/resource/$>$, dbo: $<$http://dbpedia.org/ontology/$>$, rs: $<$http://www.w3.org/2000/01/rdf-schema\#$>$

\noindent {\bf D1.} \{ ?car purl:sub\-ject db:Category:Luxury\_vehicles . ?car foaf:name ?na\-me . ?car dbo:manufacturer ?man . ?man foaf:name ?manufacturer \}

\noindent {\bf D2.} \{ ?film purl:subject db:Category:French\_films \}

\noindent {\bf D3.} \{ ?g purl:subject db:Category:First-person\_shooters . ?g foaf:name ?t \}

\noindent {\bf D4.} \{ ?p dbo:birthPlace db:Berlin . ?p dbo:birthDate ?b . ?p
purl:subject db:Category:German\_musicians . ?p foaf:name ?n . ?p rs:comment ?d\}

\noindent {\bf D5.} \{ ?per dbo:birthPlace db:Berlin . ?per dbo:birthDate ?birth . ?per foaf:name ?name . ?per dbo:deathDate ?death .\}
}
\end{minipage}
}
\caption{LUBM and DBPedia queries.}
\label{fig:queries}
\end{figure}
 \section{Conclusions}
\label{sec:conclusions}

We proposed \texttt{KOGNAC}, an algorithm for efficient encoding of RDF terms in
large Knowledge Graphs. \texttt{KOGNAC} adopts a combination of estimated
frequency-based encoding (for frequent terms) and semantic clustering (for
infrequent terms) to encode the graph efficiently and improve data locality in a
scalable way. We evaluated the performance of \texttt{KOGNAC} by integrating it
into multiple state-of-the-art RDF engines and relational stores. We observed
significant improvements regarding query runtimes, a reduction in memory usage,
and disk I/O. These results were achieved without altering the
architecture or functioning of the RDF engine, but only by rearranging the
encodings in an intelligent way.

We identified several directions for future work. More combinations of encoding,
or other clustering criteria for terms that cannot be mapped to the taxonomy
might further improve the performance. Moreover, it is interesting to study how
well our FBE and LBE encodings can deal with updates, or whether they can
improve other tasks than SPARQL. For instance, methods for knowledge completion
with embeddings might also benefit from our encodings.

To the best of our knowledge, our work is the first that seeks an improvement of
KG processing via intelligent dictionary encoding. \texttt{KOGNAC} represents a
first step in this direction to further improve the processing of emerging KGs.

{
\fakeparagraph{Acknowledgments} This work was partially funded by the
NWO VENI project 639.021.335.
\small{

\begin{thebibliography}{}

\bibitem[\protect\citeauthoryear{Angles \bgroup \em et al.\egroup
  }{2014}]{ldbc}
R.~Angles, P.~Boncz, J.~Larriba-Pey, I.~Fundulaki, T.~Neumann, O.~Erling,
  P.~Neubauer, N.~Martinez-Bazan, V.~Kotsev, and I.~Toma.
\newblock The linked data benchmark council: {A} graph and {RDF} industry
  benchmarking effort.
\newblock {\em ACM SIGMOD Record}, 43(1):27--31, 2014.

\bibitem[\protect\citeauthoryear{Bizer \bgroup \em et al.\egroup
  }{2009}]{dbpedia}
C.~Bizer, J.~Lehmann, G.~Kobilarov, S.~Auer, C.~Becker, R.~Cyganiak, and
  S.~Hellmann.
\newblock {DBpedia}-{A} crystallization point for the {Web} of {Data}.
\newblock {\em Web Semantics: Science, Services and Agents on the World Wide
  Web}, 7(3):154--165, 2009.

\bibitem[\protect\citeauthoryear{Bornea \bgroup \em et al.\egroup
  }{2013}]{rdf2db}
M.~A. Bornea, J.~Dolby, A.~Kementsietsidis, K.~Srinivas, P.~Dantressangle,
  O.~Udrea, and B.~Bhattacharjee.
\newblock {Building an Efficient RDF Store over a Relational Database}.
\newblock In {\em SIGMOD}, pages 121--132, 2013.

\bibitem[\protect\citeauthoryear{Brickley and Guha}{2014}]{rdfs}
D.~Brickley and Ramanathan~V. Guha.
\newblock {{RDF} Vocabulary Description Language 1.1: {RDF} schema}, 2014.
\newblock {W3C Recommended}.

\bibitem[\protect\citeauthoryear{Budanitsky and
  Hirst}{2006}]{budanitsky_evaluating_2006}
A.~Budanitsky and G.~Hirst.
\newblock Evaluating {WordNet}-based {measures} of {lexical} {semantic}
  {relatedness}.
\newblock {\em Comput. Linguist.}, 32(1):13--47, 2006.

\bibitem[\protect\citeauthoryear{Charikar \bgroup \em et al.\egroup
  }{2002}]{countsketch}
M.~Charikar, K.~Chen, and M.~Farach-Colton.
\newblock {Finding Frequent Items in Data Streams}.
\newblock In {\em {ICALP}}, pages 693--703, 2002.

\bibitem[\protect\citeauthoryear{Cheng \bgroup \em et al.\egroup
  }{2014}]{parallelencode}
L.~Cheng, A.~Malik, S.~Kotoulas, T.~Ward, and G.~Theodoropoulos.
\newblock Efficient parallel dictionary encoding for {RDF} data.
\newblock In {\em Proc. 17th International Workshop on the Web and Databases
  (WebDB'14)}, 2014.

\bibitem[\protect\citeauthoryear{Cormode and Muthukrishnan}{2005}]{countmin}
G.~Cormode and S.~Muthukrishnan.
\newblock {An improved data stream summary: the Count-Min Sketch and its
  applications}.
\newblock {\em J. of Algorithms}, 55(1):58--75, 2005.

\bibitem[\protect\citeauthoryear{Cur{\'e} \bgroup \em et al.\egroup
  }{2015}]{litemat}
O.~Cur{\'e}, H.~Naacke, T.~Randriamalala, and B.~Amann.
\newblock {LiteMat: {A} Scalable, Cost-efficient Inference Encoding Scheme for
  Large {RDF} Graphs}.
\newblock In {\em Big Data}, pages 1823--1830, 2015.

\bibitem[\protect\citeauthoryear{Erling and Mikhailov}{2009}]{virtuoso}
O.~Erling and I.~Mikhailov.
\newblock {Virtuoso: {{RDF}} Support in a Native {{RDBMS}}}.
\newblock In {\em Semantic Web Information Management}, pages 501--–519,
  2009.

\bibitem[\protect\citeauthoryear{Gabrilovich and
  Markovitch}{2007}]{gabrilovich_computing_2007}
E.~Gabrilovich and S.~Markovitch.
\newblock Computing {Semantic} {Relatedness} {Using} {Wikipedia}-based
  {Explicit} {Semantic} {Analysis}.
\newblock In {\em {IJCAI}}, volume~7, pages 1606--1611, 2007.

\bibitem[\protect\citeauthoryear{Gallego \bgroup \em et al.\egroup
  }{2013}]{semanticmeta}
M.~A. Gallego, O.~Corcho, J.~D. Fern{\'a}ndez, M.~A. Mart{\'i}nez-Prieto, and
  M.~C. Su{\'a}rez-Figueroa.
\newblock {\em {CAEPIA}}, chapter {Compressing Semantic Metadata for Efficient
  Multimedia Retrieval}, pages 12--21.
\newblock Springer, 2013.

\bibitem[\protect\citeauthoryear{Guo \bgroup \em et al.\egroup }{2005}]{lubm}
Y.~Guo, Z.~Pan, and J.~Heflin.
\newblock {LUBM}: {A} benchmark for {OWL} knowledge base systems.
\newblock {\em Web Semantics: Science, Services and Agents on the World Wide
  Web}, 3(2):158--182, 2005.

\bibitem[\protect\citeauthoryear{Gurajada \bgroup \em et al.\egroup
  }{2014}]{triad}
S.~Gurajada, S.~Seufert, I.~Miliaraki, and M.~Theobald.
\newblock {{TriAD}: {A} Distributed Shared-nothing {RDF} Engine based on
  Asynchronous Message Passing}.
\newblock In {\em {SIGMOD}}, pages 289--300, 2014.

\bibitem[\protect\citeauthoryear{Harbi \bgroup \em et al.\egroup
  }{2015}]{harbi2015}
R.~Harbi, I.~Abdelaziz, P.~Kalnis, and N.~Mamoulis.
\newblock Evaluating {SPARQL} queries on massive {RDF} datasets.
\newblock {\em PVLDB}, 8(12):1848--1851, 2015.

\bibitem[\protect\citeauthoryear{Harris \bgroup \em et al.\egroup
  }{2009}]{4store}
S.~Harris, N.~Lamb, and N.~Shadbolt.
\newblock {4store: The Design and Implementation of a Clustered {RDF} Store}.
\newblock In {\em {SSWS}}, pages 94--109, 2009.

\bibitem[\protect\citeauthoryear{Harris \bgroup \em et al.\egroup
  }{2013}]{sparql}
S.~Harris, A.~Seaborne, and E.~Prud'hommeaux.
\newblock {SPARQL 1.1 Query Language}.
\newblock {\em W3C Recommendation}, 21, 2013.

\bibitem[\protect\citeauthoryear{Hecht \bgroup \em et al.\egroup
  }{2012}]{seman}
B.~Hecht, S.~H. Carton, M.~Quaderi, J.~Sch\"{o}ning, M.~Raubal, D.~Gergle, and
  D.~Downey.
\newblock {Explanatory Semantic Relatedness and Explicit Spatialization for
  Exploratory Search}.
\newblock In {\em SIGIR}, pages 415--424, 2012.

\bibitem[\protect\citeauthoryear{Karp \bgroup \em et al.\egroup
  }{2003}]{frequentvalues}
R.~Karp, S.~Shenker, and C.~Papadimitriou.
\newblock A simple algorithm for finding frequent elements in streams and bags.
\newblock {\em {ACM} Transactions on Database Systems}, 28(1):51--55, 2003.

\bibitem[\protect\citeauthoryear{Karypis and Kumar}{1998}]{metis}
G.~Karypis and V.~Kumar.
\newblock A fast and high quality multilevel scheme for partitioning irregular
  graphs.
\newblock {\em SIAM Journal on Scientific Compututing}, 20(1):359--392, 1998.

\bibitem[\protect\citeauthoryear{Klyne and Carroll}{2006}]{rdf}
G.~Klyne and J.~Carroll.
\newblock {Resource Description Framework ({RDF}): Concepts and Abstract
  Syntax}, 2006.
\newblock {W3C}.

\bibitem[\protect\citeauthoryear{Kotoulas \bgroup \em et al.\egroup
  }{2010}]{skewness}
S.~Kotoulas, E.~Oren, and F.~Van~Harmelen.
\newblock {Mind the Data Skew: Distributed Inferencing by Speeddating in
  Elastic Regions}.
\newblock In {\em WWW}, pages 531--540, 2010.

\bibitem[\protect\citeauthoryear{Leal}{2013}]{semanticrel}
J.P Leal.
\newblock Using proximity to compute semantic relatedness in {RDF} graphs.
\newblock {\em Computer Science and Information Systems}, 10(4):1727--1746,
  2013.

\bibitem[\protect\citeauthoryear{Misra and Gries}{1982}]{misragries}
J.~Misra and D.~Gries.
\newblock {Finding repeated elements}.
\newblock {\em Science of Computer Programming}, 2(2):143--152, 1982.

\bibitem[\protect\citeauthoryear{Neumann and Weikum}{2008}]{rdf3x}
T.~Neumann and G.~Weikum.
\newblock {RDF-3X}: a {RISC}-style engine for {RDF}.
\newblock {\em {PVLDB}}, 1(1):647--659, 2008.

\bibitem[\protect\citeauthoryear{Pedersen \bgroup \em et al.\egroup
  }{2007}]{pedersen_measures_2007}
T.~Pedersen, Serguei V.~S. Pakhomov, S.~Patwardhan, and Christopher~G. Chute.
\newblock Measures of semantic similarity and relatedness in the {Biomedical}
  domain.
\newblock {\em Journal of Biomedical Informatics}, 40(3):288--299, 2007.

\bibitem[\protect\citeauthoryear{Robinson \bgroup \em et al.\egroup
  }{2015}]{neo4j}
I.~Robinson, J.~Webber, and E.~Eifrem.
\newblock {\em {Graph Databases: New Opportunities for Connected Data}}.
\newblock O'Reilly, 2015.

\bibitem[\protect\citeauthoryear{Sidirourgos \bgroup \em et al.\egroup
  }{2008}]{monetdb}
L.~Sidirourgos, R.~Goncalves, M.~Kersten, N.~Nes, and S.~Manegold.
\newblock {Column-Store Support for {RDF} Data Management: Not all swans are
  white}.
\newblock {\em {PVLDB}}, 1(2):1553--1563, 2008.

\bibitem[\protect\citeauthoryear{Urbani \bgroup \em et al.\egroup
  }{2013}]{compr2013}
J.~Urbani, J.~Maassen, N.~Drost, F.~Seinstra, and H.~Bal.
\newblock Scalable {RDF} data compression with {MapReduce}.
\newblock {\em Concurrency and Computation: Practice and Experience},
  25(1):24--39, 2013.

\bibitem[\protect\citeauthoryear{Urbani \bgroup \em et al.\egroup
  }{2016}]{kognacreport}
J.~Urbani, S.~Dutta, S.~Gurajada, and G.~Weikum.
\newblock {KOGNAC: Efficient Encoding of Large Knowledge Graphs (Tech.
  Report)}, 2016.
\newblock \url{http://arxiv.org/abs/1604.04795}.

\bibitem[\protect\citeauthoryear{Vrande\v{c}i\'{c} and
  Kr\"{o}tzsch}{2014}]{wikidata}
D.~Vrande\v{c}i\'{c} and M.~Kr\"{o}tzsch.
\newblock Wikidata: A free collaborative knowledge base.
\newblock {\em Commun. ACM}, 57(10), 2014.

\bibitem[\protect\citeauthoryear{Yuan \bgroup \em et al.\egroup
  }{2013}]{triplebit}
P.~Yuan, P.~Liu, B.~Wu, H.~Jin, W.~Zhang, and L.~Liu.
\newblock {{TripleBit}: A fast and compact system for large scale {RDF} data}.
\newblock {\em {PVLDB}}, 6(7):517--528, 2013.

\bibitem[\protect\citeauthoryear{Zhang \bgroup \em et al.\egroup
  }{2013}]{zhang_recent_2013}
Z.~Zhang, A.~Gentile, and F.~Ciravegna.
\newblock {Recent advances in methods of lexical semantic relatedness -- A
  survey}.
\newblock {\em Natural Language Engineering}, 19(04):411--479, 2013.

\end{thebibliography}

}}

\end{document}